\documentclass{article}
\usepackage{ijcai23}

\usepackage{times}
\usepackage{soul}
\usepackage{url}
\usepackage[hidelinks]{hyperref}
\usepackage{amsfonts,amssymb}
\usepackage[utf8]{inputenc}
\usepackage[small]{caption}
\usepackage{graphicx}
\usepackage{amsmath}
\usepackage{amsthm}
\usepackage{booktabs}
\usepackage{algorithm}
\usepackage{algorithmic}
\usepackage{color}
\usepackage{multirow}
\usepackage{float}
\usepackage{booktabs}
\usepackage[switch]{lineno}
\usepackage{pdfpages}
\usepackage{xspace}
\usepackage{enumitem}
\usepackage{makecell}
\newcommand{\ie}{\emph{i.e.,}\xspace}
\newcommand{\aka}{\emph{a.k.a.,}\xspace}
\newcommand{\eg}{\emph{e.g.,}\xspace}

\newcommand{\ignore}[1]{}
\newcommand{\tabincell}[2]{\begin{tabular}{@{}#1@{}}#2\end{tabular}}

\newcommand{\citet}[1]{\citeauthor{#1}~\shortcite{#1}}

\title{A Survey on Long Text Modeling with Transformers}

\author{
Zican Dong$^1$
\and
Tianyi Tang$^1$\and
Junyi Li$^{1,2}$\And
Wayne Xin Zhao$^{1,3}$\thanks{Corresponding author}
\affiliations
$^1$Gaoling School of Artificial Intelligence, Renmin University of China\\
$^2$DIRO, Université de Montréal\\
$^3$Beijing Key Laboratory of Big Data Management and Analysis Methods
\emails
\{210708dzc,batmanfly\}@gmail.com,\{steven\_tang,lijunyi\}@ruc.edu.cn
}

\begin{document}

\maketitle

\begin{abstract}
Modeling long texts has been an essential technique  in the field of natural language processing (NLP). With the ever-growing number of long documents, it is important to develop effective modeling methods that can process and analyze such texts.  
However, 
long texts pose important research challenges for existing text models, with  more complex semantics and special characteristics.  
In this paper, we provide an overview of the recent advances on long texts modeling based on  Transformer models. Firstly, we introduce the formal definition of long text modeling. Then, as the core content, we discuss how to process long input to satisfy the length limitation and design improved  Transformer architectures to effectively extend the maximum context length. Following this, we discuss how to adapt Transformer models to capture the special characteristics of long texts. Finally, we describe four typical applications involving long text modeling and conclude this paper with a discussion of future directions. Our survey intends to provide researchers with a synthesis and pointer to related work on long text modeling. 

\end{abstract}
\begin{figure*}[htb]
    \centering
\includegraphics[width=0.9\textwidth]{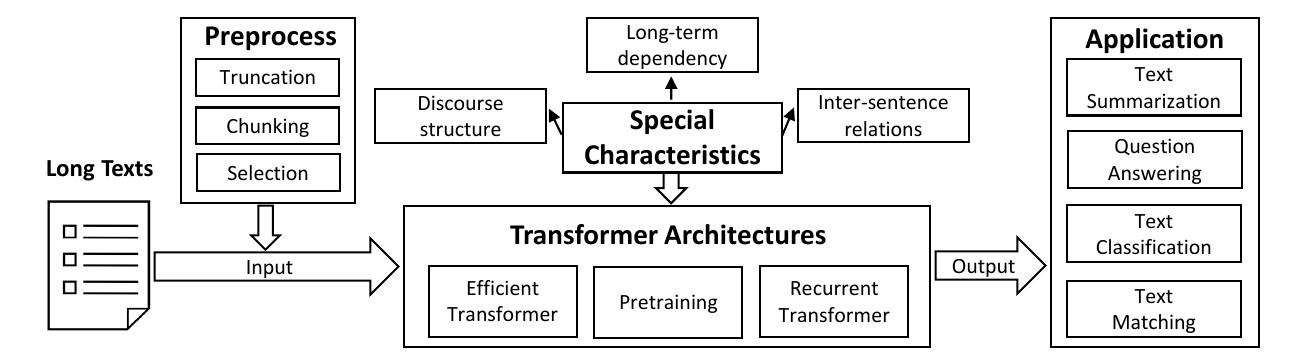}
    \caption{An illustrative process of modeling long texts with Transformers. }
    \label{fig1}
\end{figure*}

\section{Introduction}
In real life, long text is one major form of information medium that records human activities or daily events,  
\eg academic articles, official reports, and meeting scripts. 
Due to the ever-growing volume, it is difficult for humans to read, process, and extract vital and pertinent information from large-scale long texts. Consequently, there is a strong demand for NLP systems to automatically model long texts and extract information of human interest. Generally, the task of long text modeling aims to capture salient semantics from text via informative representations (\eg keywords), which would be useful for various downstream applications. For example, summarizing a long scientific paper in arXiv into its abstract~\cite{cohan-etal-2018-discourse} and classifying a long legal document into different categories~\cite{wan2019long}.


To deal with tasks involving long text modeling, many previous studies have been conducted based on recurrent neural networks (RNNs)~\cite{yang-etal-2016-hierarchical}, with the two prominent variants of LSTM~\cite{cohan-etal-2018-discourse} and GRU~\cite{yang-etal-2016-hierarchical}. Nevertheless, RNNs cannot effectively handle long-range dependencies existing in long texts. Recently, Transformer based models~\cite{vaswani2017attention}, especially the pretrained language models (PLMs), have achieved great success in NLP~\cite{devlin2018bert,radford2019language,lewis2019bart}. When it comes to employing PLMs in long text tasks, many works just adopt the same approaches to processing relatively short texts without considering the difference with long texts~\cite{lewis2019bart}.

However, modeling long texts is a challenging task in NLP. First, existing PLMs impose a \textit{length limitation} of each input sequence. Each PLM predefines a maximum context length that is usually exceeded by the length for long texts, making tokens beyond the maximum length directly discarded. Therefore, how to preprocess long texts to adapt existing PLMs is worthy of in-depth study. 
In addition, \textit{computation efficiency} is an unavoidable issue. As the length of the document increases, the time and memory consumption required to model the text increase quadratically, creating a significant burden for practical applications. 
Besides, the lengthy document contains more \textit{special characteristics} compared with the short text. Since long texts are typically domain-specific articles having complex hierarchical structures, there is a high demand for considering long-term dependency\footnote{Long-term dependency is also known as long-range dependency in some papers~\cite{pang2022long}.}, inter-sentence relations, and discourse structure. 

Despite existing works reviewing related fields, there is no survey that systematically summarizes the recent progress in long text modeling. \citet{koh2022empirical} give a brief overview of the research on long document summarization but do not go deep into the core technique of long text modeling. \citet{tay2022efficient} and~\citet{LIN2022111} focus on the topic of improving the computation efficiency of Transformer models in long text modeling. 
Different from existing reviews, this survey  attempts to provide a more general and comprehensive overview of long text modeling based on Transformer models, rather than being limited  to a specific application or topic.


The remaining sections of this survey are organized as follows (see Figure~\ref{fig1}). 
To start with, we give a formal definition of long text modeling in Section~\ref{section2}. To model long texts of arbitrary length, we introduce preprocessing methods that handle the \textit{length limitation} for PLMs in Section~\ref{section3} and Transformer architectures that efficiently extend maximum context size while keeping \textit{computation efficiency} in Section~\ref{section4}. Since long texts have \textit{special characteristics}, we explain how to design model architecture to satisfy these characteristics in Section~\ref{section5}. 
Subsequently, typical applications are presented in Section~\ref{section6}. 
Finally, we draw a conclusion from the survey and present some future directions in Section~\ref{section7}.

\section{Overview of Long Text Modeling}
\label{section2}
To begin with, we provide a formal definition of long text modeling. In this survey, a {long text} is represented as a sequence of tokens $\mathcal{X}=(x_1, \dots, x_n)$, which may contain thousands of or more tokens in contrast to short or {normal texts that can be directly processed by Transformer. 
Due to the predetermined maximum context length of PLMs, it is challenging for a Transformer model to encode an entire long sequence. Consequently, a preprocessing function $g(\cdot)$ is utilized to transform the lengthy input into a shorter sequence or a collection of snippets  (Section~\ref{section3}). 
Besides, a lengthy document will contain \textit{special characteristics} $\mathcal{C}$ 
that must be taken into account in the modeling process, \eg long-term dependency, inter-sentence relations, and discourse structure (Section~\ref{section5}). 
Further, a Transformer architecture $\mathcal{M}$ is employed to capture context information from the input data and model the semantic mapping relation from input $\mathcal{X}$ to an expected output $\mathcal{Y}$ (Section~\ref{section4}). 
Based on the concepts, the task of modeling long texts is formally described as follows:
\begin{equation}
   \label{eq1} \mathcal{Y}=f\big(g(\mathcal{X});\mathcal{C},\mathcal{M}\big). 
\end{equation}

According to the type of the output $\mathcal{Y}$, we can classify long text modeling tasks into two major categories (Section~\ref{section6}):
\begin{itemize}[leftmargin=*]
    \itemsep0em
    \item The output $\mathcal Y$ is a sequence. The setting of the task needs to capture key semantics of the long texts to obtain target sequences.
    There are two kinds of methods: the extraction method encodes the input text with an encoder and then predicts the output spans based on token-level or sentence-level representations such as question answering~\cite{gong-etal-2020-recurrent}; the generation method employs an encoder-decoder framework to generate an output text from scratch such as text summarization~\cite{cui2021sliding}. 
    \item The output $\mathcal{Y}$ is a label. The setting of the task needs to capture the crucial semantic information of the entire document for accurate classification. The input $\mathcal{X}$ is first condensed into a low-dimensional vector. Afterward, The document-level representation is fed into a classifier or employed for similarity comparison with other documents. Typical applications include text classification~\cite{park-etal-2022-efficient} and text matching~\cite{yang2020beyond}. 
\end{itemize}
\section{Preprocessing Long Input Texts}
\label{section3}
Existing Transformer-based PLMs~\cite{devlin2018bert,radford2019language,lewis2019bart} predefine the maximum sequence length, \eg  BERT can only process up to 512 tokens. 
According to Section~\ref{section2}, when the sequence length $n$ exceeds the maximum context size $t$, the preprocessing function $g(\cdot)$ is employed to transform the input document into one or more short segments (see Equation~\ref{eq1}). In this section, we introduce three major text preprocessing techniques to circumvent the \textit{length limitation} of PLMs, \ie truncation, chunking, and content selection, as summarized in Table~\ref{tab1}.

\begin{table}[htb]
    \centering
    \begin{tabular}{ll}\toprule
        Preprocessing& Formulation\\\midrule
        Text truncation & $g(\mathcal{X})=(x_1,\dots, x_t), t \le n$  \\
        Text chunking&$g(\mathcal{X})=\{\mathbf{s}_1,\dots,\mathbf{s}_l\}, \mathcal{X}=\mathbf{s}_1 \cup \cdots \cup \mathbf{s}_l$\\
         Text selection &$g(\mathcal{X})=(\mathbf{s}_{r_1},\dots,\mathbf{s}_{r_k}), k \le l$\\
         
         \bottomrule
    \end{tabular}
    \caption{Preprocessing methods and their formulation. Each snippet is a sequence of tokens $\mathbf{s_i}=(x_{i,1},\dots,x_{i,m_i})$, where $m_i$ is no larger than the maximum length $t$ of the model.}
    \label{tab1}
\end{table}
\subsection{Truncating Long Texts}

The simplest option is to truncate the long input text into a relatively brief sequence within a  predefined maximum length~\cite{lewis2019bart,park-etal-2022-efficient}, which then can be processed by an off-the-shelf PLM. In practice, the truncated text usually ranges from the beginning of the input text to the maximum length allowed by the model.

Text truncation is significantly influenced by the lead bias of the source text~\cite{zhang-etal-2022-summn}. When the crucial information is included in the segment  within the truncation length, a PLM fine-tuned with the truncated text can serve as a strong and efficient baseline, even outperforming more sophisticated models~\cite{park-etal-2022-efficient}. However, as the length increases, the salient information of a long document is typically distributed evenly throughout the entire document. Consequently, text truncation may result in significant performance degradation due to the loss of information~\cite{koh2022empirical}. In certain tasks that need to extract from the input, it would be unable to locate target spans out of truncated texts.


\subsection{Chunking Long Texts} \label{section3-chunking}
The principle of locality has been extensively observed in natural language, \ie closer texts in the document have similar semantics~\cite{liu2022leveraging}. Therefore, based on this principle, we can identify and group semantically similar segments in a long text. Given the input $\mathcal{X}$ of arbitrary length, instead of viewing it as a complete sequence, we chunk it into a sequence of snippets.
Each snippet is a sequence of tokens within the maximum length of the model, 
and the union of snippets is equal to the raw input.  Then, a PLM is employed to encode each chunk to capture the local information, followed by an aggregation operation. 


\paragraph{Designing Chunking Mechanism}
The strategy of chunking has a significant effect on the performance of downstream tasks. Commonly, the long document is divided into multiple non-overlapping parts with a fixed stride size. However, neighboring  sentences may be split into different segments, resulting in a context fragmentation problem and incomplete answers~\cite{dai-etal-2019-transformer,gong-etal-2020-recurrent}. A straightforward solution is to permit the segments to overlap~\cite{chalkidis2022exploration}. Moreover, RCM~\cite{gong-etal-2020-recurrent} proposes a reinforcement learning method to determine the stride size flexibly. When chunking lengthy and domain-specific articles, it is also essential to consider their discourse structures.
For instance, each section in a scientific paper typically covers the same topic, thus it is natural to view each section as an individual segment~\cite{liu2022leveraging}.

\paragraph{Aggregating Different Chunks}
Various methods can be utilized to aggregate the information of different chunks, depending on the downstream tasks. 
To obtain the segment-level representations, a special token (\eg \texttt{[CLS]}) is usually prepended at the beginning of each segment. Then, aggregation functions such as max-pooling and mean-pooling are employed to condense these segment representations into the document representation~\cite{li2020parade}. 
For sequence extraction tasks, the target sequence can be extracted independently from each chunk. Afterward, we select the answer with the highest score or concatenate the sequences into the final output. 
When it comes to text generation tasks, there are multiple methods to utilize the context information from each segment encoded separately. The simplest method is to concatenate the texts independently generated for each segment, which may lead to repetition issues. Based on the fusion-in-decoder module employed in open-domain QA~\cite{izacard2020leveraging}, SLED~\cite{ivgi2022efficient} encodes local context in the encoder and handles global dependency in the decoder with all effective tokens across chunks. Inspired by the principle of locality, PageSum~\cite{liu2022leveraging} makes local predictions based on individual pages and combines regional hidden states of the decoder to make final predictions.

Different from text truncation, text chunking can utilize the entire document, avoiding losing important information in the discarded text.
Nevertheless, the receptive field of PLMs is limited to a segment, resulting in the break of the long-range dependency that spans multiple segments~\cite{ding2020cogltx}. In addition, when we extract the answer from each segment independently, Since the answers are extracted from each segment independently, it is difficult to compare the scores of each segment's answer to select the best one as the final output~\cite{wang2019multi,gong-etal-2020-recurrent}.

\subsection{Selecting Salient Texts}

Based on the basic assumption that the salient information occupies only a small portion of a lengthy document and few yet important sequences can sufficiently represent the entire document~\cite{ding2020cogltx}, it is a common practice to employ a two-stage pipeline (\aka content selection), \ie identifying and concatenating relevant portions of the long text into a sequence and processing this sequence with a PLM.
As described above, the input text can be represented as a sequence of snippets. In this case, top-$k$ relevant snippets are selected and concatenated by the retriever $g(\cdot)$, without exceeding the PLM's maximum length. 


\paragraph{Designing Retrievers} 
The retriever's ability is critical to the success of selecting salient texts. Previous research has utilized unsupervised retrieval methods (\eg TextRank~\cite{mihalcea2004textrank}) that do not require training labels to enhance the simple truncation methods~\cite{park-etal-2022-efficient}. Besides, a supervised retriever can lead to additional performance improvements. LoBART~\cite{manakul-gales-2021-long} and Ext-TLM~\cite{pilault-etal-2020-extractive} employs a hierarchical RNN model to select key sentences. \citet{zhao-etal-2021-ror-read} use an existing PLM to extract fragments in conjunction with the text chunking mechanism. Inspired by the working memory mechanism of humans, CogLTX~\cite{ding2020cogltx} proposes the MemRecall procedure, which enables multi-step reasoning to extract critical text blocks via retrieval-competition and rehearsal-decay periods.

\paragraph{Training Retrievers} 
A significant challenge of content selection is to balance the training of the two stages. 
A simple method is to train the two modules in two stages separately. To train the retriever, a variety of specific methods are adopted in different situations.
For unsupervised cases in which it is difficult to obtain golden salient texts, CogLTX iteratively removes each block to adjust the relevance label based on the change of loss of the final output.
While, in certain tasks (\eg text summarization and question answering), the golden segments can be located automatically according to the target sequence~\cite{manakul-gales-2021-long,nie2022capturing}. Hence, we can train the retriever with the relevance labels in a supervised manner. To bridge the gap between data distributions of the training and inference stages, it is recommended to train the Transformer model using the union of ground-truth and extracted snippets~\cite{ding2020cogltx}. 
However, two-stage training may result in a significant disparity between the two modules, particularly for hybrid summarization tasks. To address these problems, reinforcement learning techniques are employed to optimize the entire model~\cite{bae2019summary}. Furthermore, DYLE~\cite{mao-etal-2022-dyle} assigns dynamic weights to each extracted fragment at each decoding step and proposes consistency loss to bridge the gap between the two stages.

Previous work shows that downstream tasks such as question answering can benefit from salient content selection~\cite{nie2022capturing}. 
However, the quality of the extracted texts heavily depends on the dependency on snippets throughout the long texts~\cite{zhang-etal-2022-summn}.

%





\section{Transformer Architectures for Long Texts}
\label{section4}

Considering the quadratic complexity of self-attention modules, Transformer-based PLMs do not scale well in long texts under limited computational resources. Instead of preprocessing long texts  (Section~\ref{section3}), we discuss effective architectures $\mathcal{M}$ (Equation~\ref{eq1}) for Transformer models with reduced complexity. 
Previous surveys have extensively discussed various ways to improve the \textit{computation efficiency} of Transformer~\cite{tay2022efficient,LIN2022111}. Here, we mainly discuss the variants that can efficiently extend their maximum context length. Besides, we introduce pretraining objectives and strategies of Transformers designed for long texts.



%

\subsection{Efficient Transformers}

The main consumption of time and memory comes from the self-attention mechanism in a Transformer model. A number of models are proposed to decrease the $O(n^2)$ complexity to efficiently model longer texts. 
In the rest of this section, we will discuss three major kinds of self-attention variants and how encoder-decoder attention can be improved.

\paragraph{Fixed Attention Patterns}
Based on the principle of locality of language, we can restrict attention with fixed patterns, in which each token can only attend to a few tokens according to various strategies rather than the whole sequence.
For example, the block-wise attention partitions the input text into non-overlapping blocks, and tokens can only attend to those in the same block since they are semantically similar~\cite{qiu-etal-2020-blockwise}. Similarly, local attention (\aka sliding window attention) restricts each query token to its neighbors within a fixed-size window, which can achieve $O(n)$ complexity\footnote{In fact, the complexity of local attention is $O(wn)$ where w is the length of the window.}~\cite{beltagy2020longformer,zaheer2020big}. Longformer~\cite{beltagy2020longformer} proposes a dilated attention mechanism in which there are gaps between windows, resulting in the expansion of the receptive field without additional computation. In addition, Big Bird~\cite{zaheer2020big} employs random attention, in which each token attends to a set of random tokens to enhance the model's ability to learn non-local interactions. 
Considering the \textit{long-range  dependency}, global attention is employed, where selected tokens can interact with all tokens throughout the entire sequence. We heuristically select global locations or add additional global special tokens according to task-specific requirements~\cite{beltagy2020longformer,zaheer2020big}. 
LongT5~\cite{guo-etal-2022-longt5} creates transient global attention that is constructed dynamically prior to attention operations.

\paragraph{Learnable Attention Patterns}

In addition to fixed patterns, content-based learnable attention patterns are also useful  to capture both local and global relations. The core idea is learning to assign tokens to different baskets based on the input content. Each query token can only attend to the keys in the same basket, which can improve attention efficiency. Reformer~\cite{NikitaKitaev2020ReformerTE} leverages locality-sensitive hashing to assign similar queries and keys into different hash baskets. Similarly, the K-means algorithm is used by the Routing Transformer~\cite{YiTay2020SparseSA} to cluster tokens. Sparse Sinkhorn attention~\cite{YiTay2020SparseSA} first splits a sequence into blocks and then proposes a meta sorting method to produce a query-key block assignment scheme.

\paragraph{Attention Approximation}

The preceding methods aim to select a subset of tokens for each query to focus on. As an alternative way to reduce the $O(n^2)$ complexity, we can approximate the multiplication of the query and key matrices.
In the self-attention mechanism, 
we need to calculate the similarity scores between the query and all keys, resulting in a large computation cost.
By using kernel-based approximations, 
the softmax operation can be replaced with linear dot-product of kernel feature maps $\phi(\cdot)$, \ie $\exp(\mathbf{q}_j^\top  \mathbf{k}_i) \approx \phi(\mathbf{q}_j)^\top\phi(\mathbf{k}_i)$. Afterwards, the output of self-attention can be computed as  $\mathbf{o}_j=\frac {\phi(\mathbf{q}_j)^\top\sum_{i=1}^n \phi(\mathbf{k}_i) \mathbf{v}_i^\top}{\phi(\mathbf{q}_j)^\top\sum_{i=1}^n \phi(\mathbf{k}_i)}$. Here, $\sum_{i=1}^n \phi(\mathbf{k}_i)$ and $\sum_{i=1}^n \phi(\mathbf{k}_i) \mathbf{v}_i^\top$ would  only be calculated once and reused for each query $\mathbf{q}_j$, significantly reducing the computation complexity~\cite{Transformerrnn}. In addition, different feature maps can be explored for better approximating softmax attention~\cite{Transformerrnn,HaoPeng2021RandomFA,KrzysztofChoromanski2020MaskedLM,choromanski2020rethinking}.
Considering the low-rank structure of the attention matrix, 
Linformer~\cite{SinongWang2020LinformerSW} projects keys and values into representations with a small projected length. Similarly, Luna~\cite{ma2021luna} decomposes attention into two nested operators and packs the information across the long sequence into shorter condensed tokens.

\paragraph{Efficient Encoder-decoder Attention}
Most previous work has focused on modifying the self-attention module, while the encoder-decoder attention module has been seldom discussed. As the length of the generated text increases, efficient encoder-decoder attention is increasingly vital due to the computational and memory costs. Considering the redundancy of attention heads~\cite{clark2019does,voita-etal-2019-analyzing}, HEPOs~\cite{huang-etal-2021-efficient} assign different encoder-decoder attention heads with different subsets of tokens. Similar to content selection, \citet{manakul-gales-2021-sparsity} decompose the encoder-decoder attention into two parts, \ie sentence-level attention dynamically extracts salient sentences and token-level attention only attends to tokens in the extracted subset.


\subsection{Recurrent Transformers}

The recurrent Transformer is another approach to tackling the issue of limited context length. Unlike efficient Transformers that simplify the attention structure, recurrent Transformer keeps the full self-attention mechanism. Typically, a long document is split into a series of chunks as Section~\ref{section3-chunking}. Instead of processing each chunk separately, we cache the history information of previous chunks. When the subsequent segment is fed into the model, the cached information can be exploited to mitigate the context fragmentation problem. 

-Transformer-XL~\cite{dai-etal-2019-transformer} first proposes a recurrence mechanism for Transformers. The hidden states of the previous segment are concatenated with the current segment as the input to the next layer. However, those representations of very distant segments will be discarded due to limitations of storage cost. To alleviate this problem, Compressive Transformer~\cite{rae2019compressive} condenses past representations into more coarse-grained memories. 
Due to the uni-directional self-attention, Compressive Transformer and Transformer-XL still have a fixed memory size. 
In order to eliminate this limitation, Memformer~\cite{wu2021memformer} designs a memory system with multiple slots to store history information. Memory cross attention and memory slot attention are proposed to retrieve and update the memory dynamically. ERNIE-Doc~\cite{ding-etal-2021-ernie} enhances recurrent mechanisms through the same-layer recurrence, \ie passing the concatenation of the hidden states of the current segment and the previous segment at the next layer to the next layer, further extending the context length while maintaining fine-grained representations. In addition, a retrospective feed mechanism is applied,  making the global context information of a document available for each segment.

\subsection{Pretraining for Long Texts}

Recently, pretraining has been proven to be an effective way to learn task-agnostic representations for improving performance on various NLP tasks, especially in the data-scarcity setting. Typically, PLMs are pretrained with texts that are typically shorter than the target data.
Due to the discrepancy between short and long texts, different pretraining configurations should be explored for the lengthy sequences. Considering that pretraining on long documents from scratch is expensive, we can continue pretraining based on existing PLMs checkpoints~\cite{beltagy2020longformer,zaheer2020big}.


The pretraining tasks are essential for Transformer models to acquire task-agnostic representations from data. Generally, PLMs adopt standard language modeling~\cite{radford2019language} and denoising autoencoding objectives~\cite{devlin2018bert,lewis2019bart}. 
To pretrain on long documents, the majority of models adopt masked language modeling (MLM) as their objectives, such as Longformer~\cite{beltagy2020longformer} and Big Bird~\cite{zaheer2020big}. Since the MLM task mainly focuses on token-level semantics, more pretraining objectives have been explored for sentence-level or document-level relations. HIBERT~\cite{zhang-etal-2019-hibert} randomly masks the whole sentences from long documents and employs the sentence prediction task for better capturing inter-sentence relations. Similarly, PEGASUS~\cite{zhang2020pegasus} proposes a gap sentences generation (GSG) task, where top-k
sentences selected based on the ROUGE1 scores compared to the rest of the document are masked to be reconstructed. Despite being designed for text summarization tasks, the GSG task has been shown effective for different long text tasks~\cite{guo-etal-2022-longt5,phang2022investigating}. Additionally, ERNIE-Doc~\cite{ding-etal-2021-ernie} introduces the document-aware segment-reordering objective to reorganize the permuted segments.




\section{Special Characteristics of Long Text}
\label{section5}
By utilizing the techniques described in Sections~\ref{section3} and~\ref{section4}, it is possible to model a long document in an effective manner. At the same time, compared to short texts, long documents have more \textit{special characteristics} $\mathcal {C}$ (Equation~\ref{eq1}) that must be taken into account. As shown in Table~\ref{tab2}, we mainly consider three typical characteristics: long-term dependency, inter-sentence relation, and discourse structure, and discuss how to enforce  these characteristics when modeling long texts.

\begin{table*}

\centering
\small
\begin{tabular}{c|c|l}
\toprule 
Characteristics &Methods& Related Works \\
\midrule 
\multirow{4}{*}{\tabincell{c} {long-term\\ dependency}}& \tabincell{c}{Enhancing\\ local attention} &\tabincell{l}{Global attention~\cite{beltagy2020longformer,zaheer2020big},\\segment-level representations~\cite{pang2022long}, and state space models~\cite{zuo2022efficient}.}\\\cmidrule{2-3}
& \tabincell{c}{Modeling cross\\segment interaction} &\tabincell{l}{Recurrence mechanism~\cite{gong-etal-2020-recurrent}, memory module~\cite{cui2021sliding},\\cross-segment encoder~\cite{pappagari2019hierarchical}, and multi-stage framework~\cite{zhang-etal-2022-summn}}\\\midrule
\multirow{4}{*}{\tabincell{c} {Inter-sentence\\Relations}}&\tabincell{c} {Hierarchy-based\\models}&\tabincell{l} { Inter-sentence encoder layers~\cite{zhang-etal-2019-hibert,ruan-etal-2022-histruct,cho2022toward}\\and additional sentence-level cross-attention module~\cite{rohde2021hierarchical}}\\\cmidrule{2-3}
&\tabincell{c}{Graph-based\\models}&\tabincell{l}{Graphs with hierarchical structures~\cite{phan-etal-2022-hetergraphlongsum},\\and topic enhanced graphs~\cite{cui-etal-2020-enhancing,zhang2022hegel}}\\\midrule

\multirow{4}{*}{\tabincell{c}{Discourse\\Structure}}&\tabincell{c}{Explicitly\\designing models}&\tabincell{l}{Hypergraphs with section hyperedges~\cite{zhang2022hegel},\\section title embedding, and section position embedding~\cite{ruan-etal-2022-histruct}}\\\cmidrule{2-3}
&\tabincell{c}{Implicitly\\enhancing models}&\tabincell{l}{Section segmentation objective~\cite{cho2022toward}\\and chunking according to the sections~\cite{liu2022leveraging}}
\\

\bottomrule    
\end{tabular}
\caption{Categories of special characteristics of long texts.}
    \label{tab2}
\end{table*}

\subsection{Long-term Dependency}
long-term dependency is an important property of long documents, where distant tokens may be related to each other in the semantics. Equipped with the self-attention mechanism, Transformer is more capable of  modeling  long-term dependency than recurrent neural networks~\cite{vaswani2017attention}. Nonetheless, lengthy documents are typically divided into disjoint segments or encoded with the sparse attention mechanism, jeopardizing long-distance semantic relations. To tackle this issue, there are two major methods to preserve long-term dependency on long documents.

\paragraph{Enhancing Local Attention}
Considering the \textit{computation efficiency} requirement, local attention is typically substituted for full self-attention when modeling long texts~\cite{beltagy2020longformer,zaheer2020big,guo-etal-2022-longt5,ainslie-etal-2020-etc}. With a small receptive field, the model with local attention has a limited ability for capturing long-term dependencies.Global attention mentioned in Section~\ref{section4} is an effective method of enhancing local attention~\cite{beltagy2020longformer,zaheer2020big}. Further, Top Down Transformer~\cite{pang2022long} stacks the segment-level Transformer layers on top of the Transformers with local attention to obtain segment-level representations. Then, the segment-level outputs are used to enhance token representations through cross-attention. SAPADE~\cite{zuo2022efficient} employs state space models tailored for long input before local self-attention layers to capture long-term dependency.

\paragraph{{Modeling Cross-segment Interaction}}
Independently encoding each segment cannot well capture long-distance interactions across different segments since they can only aggregate local information.
. To resolve this issue, we can enable unidirectional or bidirectional information flow across segments via additional modules.
RCM~\cite{gong-etal-2020-recurrent} and RoBERT~\cite{pappagari2019hierarchical} encode each snippet into a low-dimensional vector to transfer information between segments with a recurrent function such as an LSTM and a gated function.
Recurrent Transformers modify Transformer architectures to transmit both fine-grained and coarse-grained memories of preceding representations into current window~\cite{dai-etal-2019-transformer,rae2019compressive}. 
Another way to model global information is the memory network that can preserve past segments' information. In SSN-DM~\cite{cui2021sliding}, salient information is maintained through a dynamic memory module and interacts with the current segment using graph neural networks (GNNs). Similarly, CGSN~\cite{nie2022capturing} utilizes a hierarchical global graph to preserve and dynamically update multi-level historical information.
Besides the unidirectional approaches mentioned above, we can also model the bidirectional interaction across windows. Normally, we stack segment-level representations into a sequence. Then, this sequence is processed by a cross-segment encoder (\eg Transformer and BiGRU) that may be stacked on the top of the segment-wise Transformer or inserted into each Transformer layer. 
~\cite{pappagari2019hierarchical,grail2021globalizing}. 
SUMM$^N$~\cite{zhang-etal-2022-summn} proposes a multi-stage split-then-summarize framework in which the concatenation of each segment's local summary serves as the input to the subsequent stage. Therefore, in the final step, the entire receptive field is ensured in the brief input.


\subsection{Inter-sentence Relations}
A long document usually consists of multiple paragraphs and sentences. For existing Transformer models, a special token such as \texttt{[CLS]} is typically inserted at the beginning of each sentence to represent a sentence. However, the vast majority of Transformer models are pretrained to learn token-level representations, which cannot well capture cross-sentence dependency. In this case, it is insufficient for the representation of the special token to aggregate complex intra-sentence and inter-sentence information. There are two primary types of methods for capturing inter-sentence relations: graph-based and hierarchy-based models.

\paragraph{Hierarchy-based Models}
From top to bottom, a long document can be divided into paragraphs, sentences, and tokens. In order to utilize the hierarchical information containing complex inter-sentence dependencies, Transformer architectures can be modified in a hierarchical manner.
Inspired by HAN~\cite{yang-etal-2016-hierarchical} that designs hierarchical RNN modules, multiple studies propose stacking inter-sentence Transformer layers on top of off-the-shelf pretrained Transformer encoders~\cite{zhang-etal-2019-hibert,ruan-etal-2022-histruct,cho2022toward}. As for Transformer encoder-decoder models, HAT~\cite{rohde2021hierarchical} and HMNet~\cite{zhu-etal-2020-hierarchical} inserts an additional cross-attention module that attends over the hidden states encoded by the sentence-level Transformer.

\paragraph{Graph-based Models}
Graph-based model is another approach to modeling high-order dependency. Generally, a lengthy document can be transformed into a graph where semantic units are represented as nodes and intricate relationships are modeled as the edges. Therefore, downstream tasks can be converted  into a node classification problem. In this setting, the Transformer model serves as an encoder to initialize the hidden states of each node. The majority of work constructs the graph at the sentence level. Latent topic models are adopted to extract a variety of topics in a document. Each topic has its own representations and these topic representations are connected with sentence representations through edges~\cite{cui-etal-2020-enhancing,zhang2022hegel}. 
Additionally, heterogeneous graphs with hierarchical structures can enrich the node representations with the word- and passage-level information~\cite{phan-etal-2022-hetergraphlongsum,doan-etal-2022-multi}. After constructing the graph, a graph neural network is utilized to update the hidden states of nodes by aggregating the information of neighbors to learn various cross-sentence relations, and these hidden states are fed into a classifier for ranking or predicting confidence scores.

\subsection{Discourse Structure}
Discourse is defined as the semantic unit containing multiple sentences in the NLP field. In contrast to short texts, long documents (\eg scientific articles and books) usually have complex discourse structures, \eg sections, and paragraphs.
By leveraging such structural information, models can boost their performance on downstream tasks. 

On the one hand, it is possible to explicitly design models to inject discourse structure information. Typically, a scientific article consists of several sections that contain multiple sentences, and different sections focus on different semantic topics.
Based on this property, HEGEL~\cite{zhang2022hegel} creates hyper-edges to connect all of the sentences within a section. Hierarchical position embedding is designed by HIBRIDS~\cite{cao-wang-2022-hibrids} and HiStruct+~\cite{ruan-etal-2022-histruct} to better capture the section-level discourse structure. Furthermore, the title of each section that highly summarizes the theme of the segment should be taken into account~\cite{ruan-etal-2022-histruct}.

On the other hand, this characteristic can also be exploited implicitly. To enhance performance on long document summarization, an additional section segmentation task, \ie predicting section boundaries, is proposed in Lodoss~\cite{cho2022toward}. Additionally, we can exploit the inductive bias of discourse structure when chunking the document into multiple segments. Each section will be viewed as a separate segment~\cite{liu2022leveraging,manakul-gales-2021-long}.
\section{Application}
\label{section6}
The technique of long text modeling is widely useful in a variety of downstream applications. 
In this part, we mainly discuss four typical applications, \ie text summarization, question answering, text classification, and text matching. 

\paragraph{Text Summarization}
Text summarization is an application of condensing a source text into a brief text while retaining the essential details. 
When summarizing long documents in an extractive way, content redundancy tends to occur, \ie segments with similar meanings are extracted.  Graph-based and hierarchy-based methods are employed due to their ability to reduce the redundancy between extracted sentences~\cite{phan-etal-2022-hetergraphlongsum,cho2022toward}.
In addition, instead of directly computing the confidence scores, \citet{xiao2020systematically} explore more complex methods to balance redundancy and importance. In abstractive summarization, local attention or chunking mechanism are utilized to capture the semantic information of long documents~\cite{guo-etal-2022-longt5,liu2022leveraging}. Based on the information, a fluent, coherent summary can be generated. Besides, the two methods above can be combined into hybrid summarization, \ie 
 an extraction-then-generation pipeline~\cite{manakul-gales-2021-long,mao-etal-2022-dyle,pilault-etal-2020-extractive}.

\paragraph{Question Answering}
Long document question answering aims to understand the source document and extract or generate answers given a question. Besides the methods that directly extract answers, most works follow a retrieve-then-read process, \ie we first select relevant evidence and then apply a PLM to extract spans or generate the answer on the brief evidence~\cite{nie2022capturing}. As for the multi-hop QA tasks, the reasoning process spans multiple articles, and these articles are concatenated into a long document. Then, multi-step reasoning iteratively retrieves and updates the answer based on the question~\cite{ding2020cogltx}.

\paragraph{Text Classification}
Text classification aims to assign  texts into a number of pre-defined  categories, with applications including sentiment analysis, natural language inference, and news classification. For PLMs, the embedding of the special token (\eg \texttt{[CLS]}) is taken as the representation of the entire document and fed into a softmax function for predicting the classification label.  In long document classification, fine-tuning a BERT after truncating long texts can be a competitive baseline~\cite{park-etal-2022-efficient}. Besides this method, efficient attention, chunking, and selection mechanism have been employed to enhance the semantic modeling by preprocessing more tokens~\cite{zhang-etal-2019-hibert,beltagy2020longformer,ding2020cogltx}. While, \citet{park-etal-2022-efficient} systematically evaluate different models and conclude that complex models do not consistently work well across datasets, demonstrating the need for advanced models with robust performance.

\paragraph{Text Matching}

Text matching aims to estimate the semantic relation between a source and target text pair according to their similarity. Semantic matching between long document pairs has a variety of applications, such as article recommendation and citation recommendation, but it is more challenging due to the need for a comprehensive understanding of  semantic relations~\cite{10.1145/3308558.3313707,yang2020beyond}. 
In existing work, a two-tower Siamese network structure~\cite{bromley1993signature} is widely used by  calculating  the similarity between the representations of two documents.
To condense the long document into a low-dimensional vector, 
similar approaches for long text classification can be also used. For example, SMITH~\cite{yang2020beyond} chunks a long document into several segments and aggregates information from different segments into the document-level representation through inter-segment Transformer layers.

\section{Conclusion and Future Directions}
\label{section7}
In this paper, we provide an overview of the recent advances in long text modeling based on Transformer models. We first summarize the preprocessing techniques for long input texts, efficient Transformer architectures for reducing the computation complexity, and special characteristics in long texts. Furthermore, four typical applications based on long text modeling  are discussed. We conclude this paper by presenting several future directions for long text modeling:


\paragraph{Exploring Architectures for Long Texts}

The Transformer model has achieved remarkable success in NLP, but the quadratic computation complexity limits its application to long texts. Even though there are different variants that can efficiently model long texts, their performance still generally underperforms the full self-attention mechanism.  Therefore, it is worthwhile to explore more efficient architectures for modeling long texts. Recently, state space models~(SSMs) have shown their superiority in multiple long sequence modeling tasks.  SPADE~\cite{zuo2022efficient} enhances local attention with SSMs, greatly outperforming existing models with only linear complexity. In the meantime, all-MLP architectures 
have been shown to be cost-effectively competitive with Transformers~\cite{liu2021pay}, and we can further explore their potential for modeling long text domains. 

\paragraph{Designing Long-text PLMs}
Transformer models have been widely adopted for long text tasks, and most of them adopt the MLM objective or its variants with different masking techniques~\cite{devlin2018bert,zhang-etal-2019-hibert}. ERNIE-Doc proposes coarser-grained pretraining methods (\eg document-aware segment-reordering objective) to capture high-level semantics, which can be considered a powerful complement to MLM. Moreover, task-specific pretraining objectives demonstrate their effectiveness in various long document tasks~\cite{zhang2020pegasus,guo-etal-2022-longt5}. In the future, it will be promising to investigate how to design advanced pretraining methods and develop more powerful models specially for long texts.

\paragraph{Eliminating Gaps between Existing PLMs and Long Texts} Reusing existing PLMs in long texts is a promising method but the majority of them are not designed for long texts~\cite{gong-etal-2020-recurrent,ivgi2022efficient,zhang-etal-2022-summn}. As discussed in Section~\ref{section3}, there are mainly three types of approaches to adapting existing PLMs to long texts, \ie text truncation, text chunking, and text selection. Although these methods can transform long texts into sequences that can be directly processed by the PLMs, there still exists a gap between the processed segments and the pretraining texts.
In addition, salient information may be lost and dependencies may be broken during the stage of preprocessing. Consequently, future research can investigate how to eliminate the mismatch between the model and data and how to preserve information to the greatest extent possible.


\paragraph{Modeling in Low-resource Settings}
There is a great need for data to train a Transformer due to its large number of parameters. However, Low-resource settings are common in real practice where labeled data for supervised training is scarce.
To overcome the challenge, efficient tuning techniques can be employed to bridge the gap between pretraining and finetuning while avoiding overfitting~\cite{pu-etal-2022-two}. 
Besides, data augmentation is another effective method  for long text summarization tasks. Se3~\cite{moro2022semantic} splits a long document into multiple segments with the corresponding targets for separate summarization, creating more data instances to alleviate the data scarcity problem. Additionally, \citet{bajaj-etal-2021-long} extract salient snippets and transfer the knowledge from a BART continually pretrained on 
datasets on a related domain.
Since manually collecting and processing long texts are time-consuming, more effort is required to process long texts in low-resource situations.

\paragraph{Modeling with Large Language Models}
Recently, large language models (LLMs) have achieved remarkable performance in various NLP tasks and satisfy human preferences due to their strong zero-shot or few-shot generalization capabilities. Typically, the maximum lengths of the context window in LLMs greatly exceed those in traditional PLMs, (\eg 2048 tokens in GPT-3~\cite{brown2020language} and 4000 tokens in InstructGPT~\cite{ouyang2022training}), making it possible for LLMs to directly process the long document. Meanwhile, we can directly employ LLMs to handle long text tasks through task instructions or a few exemplars without fine-tuning the model, significantly reducing the computation consumption. However, there still exist texts exceeding the maximum length and the length of each exemplar is limited by the total length in the few-shot settings. Therefore, it is still worth exploring how to process long texts with LLMs to deal with the length limitation and satisfy special characteristics. As a pioneer work in applying LLM to long text problems, PCW~\cite{ratner2022parallel} chunks the lengthy input and separately encodes each window with an LLM. Then, all windows will be attended to during the generation stage using the fusion-in-decoder mechanism~\cite{izacard2020leveraging}.

\bibliographystyle{named}
\bibliography{ijcai23}

\end{document}